\pdfoutput=1

\documentclass[11pt]{article}

\usepackage{acl}

\usepackage{times}
\usepackage{latexsym}

\usepackage[T1]{fontenc}

\usepackage[utf8]{inputenc}

\usepackage{microtype}

\usepackage{graphicx}
\usepackage{subfigure}
\usepackage{amsmath}

\title{Thinking with DistilQwen: A Tale of Four Distilled Reasoning and Reward Model Series}

\author{Wenrui Cai$^{1,2}$\thanks{\ \ The work was conducted during the internship at Alibaba Cloud Computing.}, Chengyu Wang$^2$\thanks{\ \ Corresponding author.}, Junbing Yan$^2$, Jun Huang$^2$, Xiangzhong Fang$^1$\\
  $^1$ Shanghai Jiao Tong University, Shanghai, China\\
  $^2$ Alibaba Cloud Computing, Hangzhou, China\\
  \texttt{\{cwrcwr,xzfang\}@sjtu.edu.cn}\\
  \texttt{\{chengyu.wcy,yanjunbing.yjb,huangjun.hj\}@alibaba-inc.com}}

\begin{document}
\maketitle
\begin{abstract}
Recently, the demand for small and efficient reasoning models to support real-world applications has driven the development of knowledge distillation techniques that balance reasoning performance and inference speed. In this paper, we further extend the DistilQwen model family, initialized from the Qwen models, by introducing four model series specifically designed to meet industrial requirements. The distilled model collection comprises: 
(1) slow-thinking models, optimized for reasoning tasks that require high accuracy; 
(2) two series of adaptive-thinking models, which dynamically adjust reasoning strategies based on input tasks to maximize efficiency across diverse scenarios; and 
(3) distilled reward models, which enable further reinforcement learning of reasoning models using distilled knowledge.
Comprehensive evaluations across multiple benchmarks demonstrate both high inference efficiency and strong reasoning performance for these models, as well as the practical utility of distilled reward models. We further show that these models support industry practitioners by providing scalable training and inference functionalities on the Alibaba Cloud PAI (Platform for Artificial Intelligence) platform.\footnote{Resources are released in the EasyDistill toolkit~\cite{DBLP:journals/corr/abs-2505-20888}. URL: \url{https://github.com/modelscope/easydistill}}
\end{abstract}

\section{Introduction}

In the rapidly evolving landscape of large language models (LLMs), the need for efficient reasoning models that can seamlessly integrate into real-world applications has become increasingly urgent. Industries worldwide increasingly rely on advanced LLMs, which require not only high reasoning performance but also fast inference speeds to support timely decision-making~\cite{DBLP:journals/cbm/VrdoljakBMSKOCBV25,DBLP:journals/kbs/ZhongHWLY25,DBLP:conf/chi/ChaPCRS25}. These dual requirements have fueled growing interest in knowledge distillation (KD) methods, which aim to balance accuracy with computational efficiency~\cite{DBLP:journals/corr/abs-2402-13116}.

\begin{figure}[t]
    \centering
    \includegraphics[width=.485\textwidth]{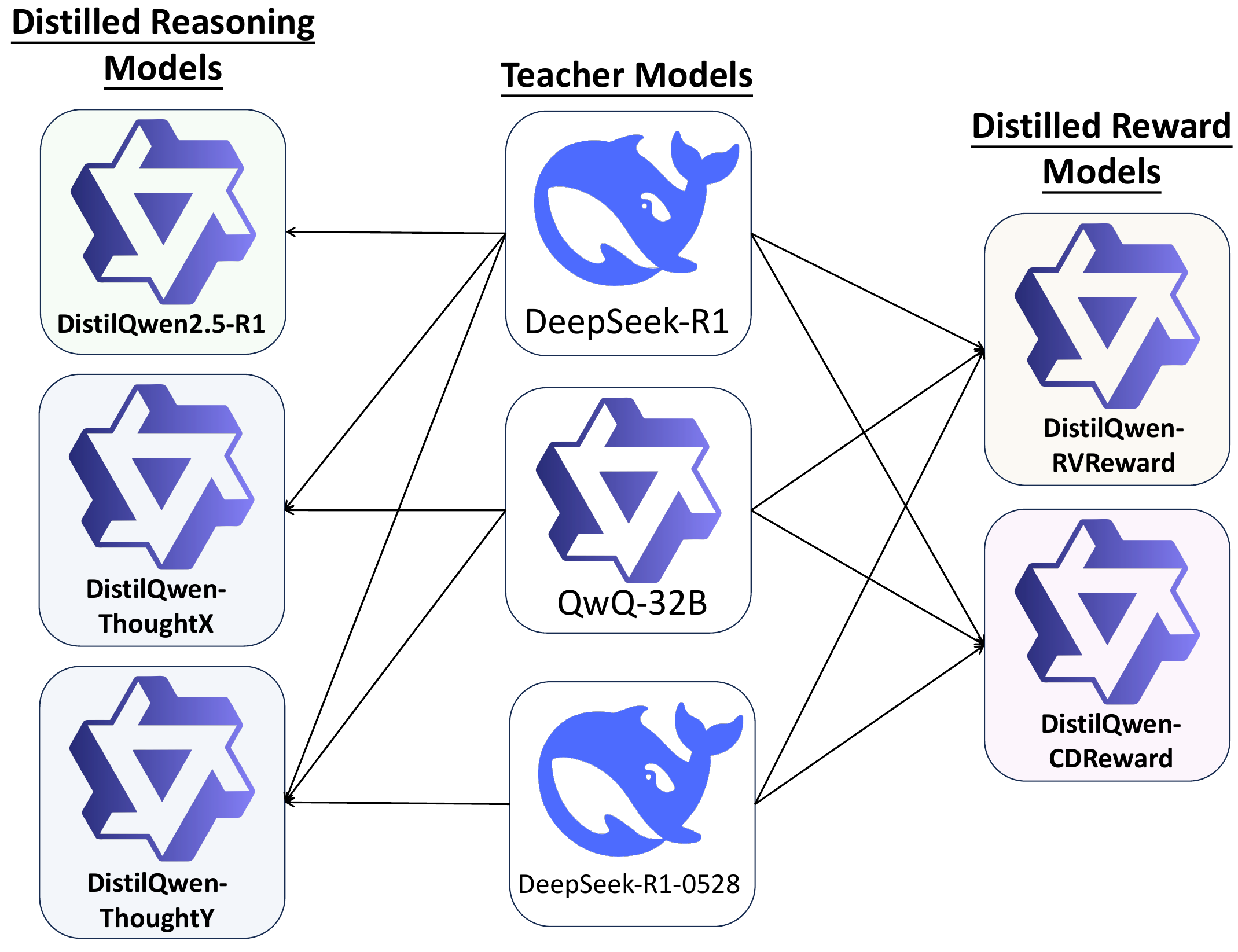}
    \caption{Roadmap for training DistilQwen reasoning and reward models.}
    \label{fig:roadmap}
\end{figure}

\begin{figure*}[t]
    \centering
    \includegraphics[width=\textwidth]{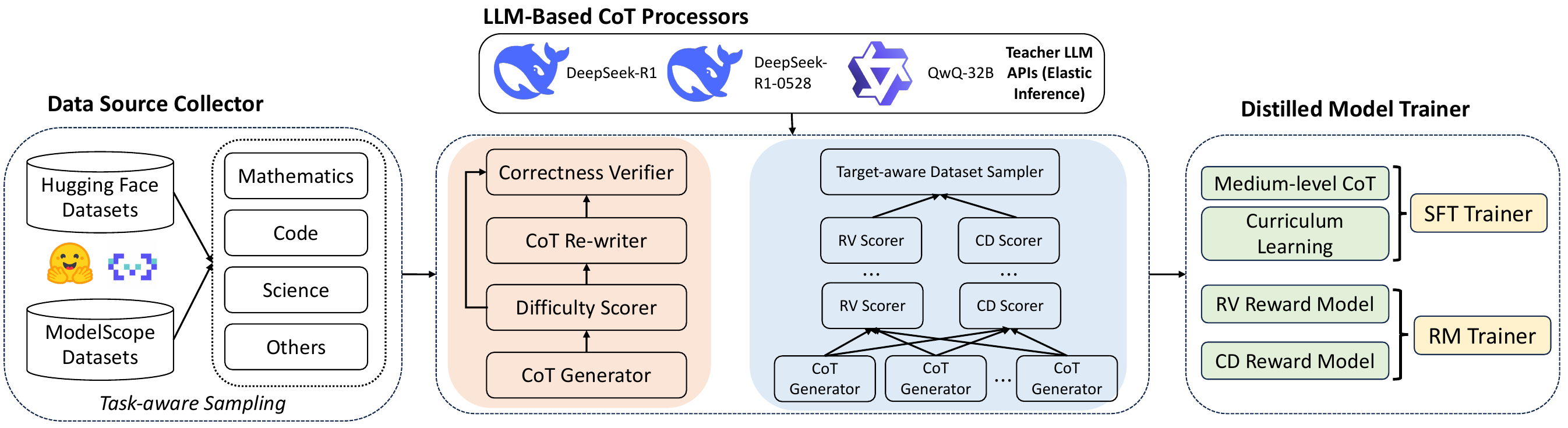}
    \caption{High-level process for obtaining DistilQwen reasoning and reward models.}
    \label{fig:process}
\end{figure*}

In response to these industrial needs, we present a comprehensive extension of the DistilQwen model family~\cite{DBLP:journals/corr/abs-2504-15027}, introducing four model series tailored for a wide array of reasoning scenarios:

\begin{itemize}
    \item \textbf{Slow-thinking models:} Optimized for tasks where accuracy is paramount, these models ensure consistently high precision.
    
    \item \textbf{Adaptive-thinking models (two series):} Motivated by recent insights that reasoning length and depth should vary by task~\cite{DBLP:journals/corr/abs-2503-16419,DBLP:journals/corr/abs-2504-09802}, these models dynamically adjust their reasoning strategies to specific requirements and outperform previous ones.
    
    \item \textbf{Adaptive-thinking-based reward models:} Derived from our training strategies for above series, these models support further reinforcement learning using distilled knowledge.
\end{itemize}

The development roadmap for these model series is shown in Figure~\ref{fig:roadmap}.
DistilQwen reasoning and reward models are evaluated on diverse benchmarks, demonstrating robust performance. Furthermore, practicality is demonstrated by their integration into industrial AI platforms for fine-tuning and online deployment.

\section{Related Work}

Knowledge distillation (KD), first introduced by \citet{DBLP:journals/corr/HintonVD15}, has been key to reducing parameter counts in language models. Before LLMs, several studies distilled BERT-based models~\cite{DBLP:journals/corr/abs-1910-01108,DBLP:conf/emnlp/JiaoYSJCL0L20,DBLP:conf/acl/SunYSLYZ20,DBLP:conf/acl/Pan0QZLH20}, mainly for natural language understanding. Distillation for LLMs poses additional challenges due to complex token dependencies. 

In the literature, MiniLLM~\cite{DBLP:conf/iclr/Gu0WH24} uses a reverse Kullback-Leibler divergence (KLD) objective to transfer knowledge from white-box LLMs to student models.
An adaptive strategy~\cite{DBLP:conf/coling/WuTWY0W25} combines forward and reverse KLD via dynamic weighting.
$f$-Distill~\cite{DBLP:conf/acl/Wen0DM23} minimizes a generalized $f$-divergence function at the sequence level.
\citet{DBLP:conf/coling/LiZS25} propose a bi-directional logits difference loss to improve KD performance.
For black-box KD (where only APIs are available), knowledge distillation signals like output logits are absent; researchers leverage data augmentation with instruction-response pairs~\cite{DBLP:conf/acl/HsiehLYNFRKLP23,DBLP:conf/acl/LiCCHGZ24,DBLP:conf/iclr/Lou0XSAX0024,DBLP:conf/emnlp/RanaldiF24,DBLP:conf/naacl/KimJK24,DBLP:conf/emnlp/YueWHW24,DBLP:conf/aaai/MaCZ0D25}.

With the rise of large reasoning models such as DeepSeek-R1~\cite{DBLP:journals/corr/abs-2501-12948}, distilling Chain-of-Thought (CoT) reasoning ability has attracted attention~\cite{DBLP:journals/corr/abs-2504-09100}.
\citet{DBLP:journals/corr/abs-2502-18001} study the effect of granularity, format, and teacher choice on CoT distillation.
\citet{DBLP:journals/corr/abs-2504-09802} improve reasoning in small models by considering cognitive gaps.
Multi-teacher KD~\cite{DBLP:conf/wsdm/0001HC0C25} uses teacher forcing to transmit diverse reasoning skills.
Self-training methods~\cite{DBLP:journals/corr/abs-2502-12744} activate latent reasoning capacity in small models.

In our work, we open-source a range of small models for challenging reasoning tasks to facilitate further research and industrial applications.

\section{Algorithm Implementation}

We elaborate on our industrial practice for training the DistilQwen reasoning and reward models. The overall pipeline is illustrated in Figure~\ref{fig:process}.

\subsection{Data Source Collector}

The foundation of our approach is the Data Source Collector, which aggregates CoT datasets from platforms such as Hugging Face\footnote{\url{https://huggingface.co/datasets}} and ModelScope\footnote{\url{https://modelscope.cn/datasets}}. These datasets span domains including mathematics, code, science, and more, providing rich and diverse sources for training reasoning models such as OpenThoughts2\footnote{\url{https://huggingface.co/datasets/open-thoughts/OpenThoughts2-1M}}, DeepMath-103K~\cite{DBLP:journals/corr/abs-2504-11456}, OpenCodeReasoning\footnote{\url{https://huggingface.co/datasets/nvidia/OpenCodeReasoning}}, etc. We subsequently perform task-aware re-sampling to balance the distributions across different types of tasks.

\subsection{LLM-Based CoT Processors}

Directly performing vanilla SFT training on raw CoT datasets does not necessarily yield strong student models. Below, we describe our LLM-based CoT processors, which effectively and efficiently generate and refine CoT datasets for knowledge distillation (KD).

\subsubsection{Elastic Teacher LLM Inference}
In our implementation, directly invoking third-party APIs for teacher LLM inference is not feasible due to the requirements for robust, scalable, and elastic inference. Instead, we deploy inference services for DeepSeek-R1~\cite{DBLP:journals/corr/abs-2501-12948}, DeepSeek-R1-0528\footnote{\url{https://huggingface.co/deepseek-ai/DeepSeek-R1-0528}}, and QwQ-32B\footnote{\url{https://huggingface.co/Qwen/QwQ-32B}} on our computing clusters, where each server is equipped with eight NVIDIA H20 GPUs (96GB each). Because the queries per second (QPS) requirements for these models vary across subsequent steps, the number of inference nodes per model can be elastically adjusted to maximize computational resource utilization. These inference APIs form the foundation of our system; thus, we do not need to manage hardware for CoT generation and processing later.

\subsubsection{Slow-Thinking CoT Processor}

The slow-thinking CoT processor is employed to optimize CoT training sets for \textbf{slow-thinking models}, i.e., the \textbf{DistilQwen2.5-R1} series (using DeepSeek-R1~\cite{DBLP:journals/corr/abs-2501-12948} as the teacher model and the Qwen2.5 series as students).

\noindent\textbf{CoT Generator.}
The core module is the CoT Generator, which leverages DeepSeek-R1 to generate structured CoTs by exploring complex solution spaces. However, the output reflects how this ultra-large model (with 671B parameters) solves problems, which may not be entirely suitable for smaller models to learn from. To create diverse solution paths, multiple inference outcomes are generated using varying temperatures.

\noindent\textbf{CoT Difficulty Scorer.}
A key challenge in improving KD effectiveness for smaller student models is addressing the capacity gap between teacher and student models. Several concurrent works~\cite{DBLP:journals/corr/abs-2504-09802,DBLP:journals/corr/abs-2504-11919} propose assessing and rewriting CoTs to better suit student learning; however, these methods often require iterative processing and incur high computational costs. In our approach, the CoT Difficulty Scorer evaluates the complexity of each generated CoT using the same teacher LLM (DeepSeek-R1). By assigning difficulty levels (easy, medium, hard), the scorer distinguishes intricate CoTs from simpler or excessively challenging ones, enabling students to prioritize learning on medium-level CoTs. This systematic scoring helps models develop a deeper understanding of complex reasoning scenarios.

\noindent\textbf{CoT Re-writer and Verifier.}
As reported by~\citet{DBLP:journals/corr/abs-2504-09802}, rewritten and verified versions of CoT datasets often enable smaller LLMs to achieve stronger reasoning abilities. However, the complexity of these processing steps limits parallelization over millions of CoTs in our dataset. Therefore, we adopt a simple yet effective strategy by rewriting and verifying CoTs only at the easy and hard difficulty levels in a one-pass process. Incorrect CoTs are discarded based on the verifier, which serves as a safeguard to preserve reasoning correctness. Since multiple CoTs are generated per problem, in most cases, at least one suitable CoT is obtained. Overall, integrating these steps results in large, higher-quality CoT training sets that constitute the foundation for training our \textbf{slow-thinking models}.

\subsubsection{Adaptive-Thinking CoT Processor}

Beyond the slow-thinking processor, the adaptive-thinking CoT processor further optimizes CoT training sets for \textbf{adaptive-thinking models}, namely the \textbf{DistilQwen-ThoughtX} series (using DeepSeek-R1 and QwQ-32B as teacher models and the Qwen2.5 series as students), and its updated version following the release of DeepSeek-R1-0528: the \textbf{DistilQwen-ThoughtY} series (using DeepSeek-R1, DeepSeek-R1-0528, and QwQ-32B as teacher models and the Qwen3 series\footnote{\url{https://qwenlm.github.io/blog/qwen3/}} as students). These models dynamically adjust the lengths of CoTs according to problem complexity, thereby further improving reasoning abilities and avoiding ``over-thinking.''

\noindent\textbf{CoT Generator.}
The implementation of the CoT Generator here is largely similar to that of the slow-thinking processor. The key difference is that we generate multiple CoTs per problem using different teacher models. Additionally, the inference temperatures are varied to increase diversity.

\noindent\textbf{RV and CD Scorers.}
During the upgrade of DistilQwen reasoning models, we discovered that rewriting alone is insufficient, as it does not address situations where models tend to ``over-think'' simpler problems~\cite{DBLP:journals/corr/abs-2503-16419}. Our Reasoning Verbosity (RV) and Cognitive Difficulty (CD) Scorers, derived from the work in~\cite{omnithought}, further assess the quality of CoTs by ensuring they are appropriately verbose for challenging problems and match the cognitive capacity of the student models. This dual scoring mechanism ensures that models are exposed to a broad spectrum of CoT processes that are better aligned with their capabilities and the problem difficulty. Consequently, the models can learn to adaptively think based on the input problems, leading to higher accuracy and faster inference.

\noindent\textbf{Target-aware Dataset Sampler.}
Finally, given a target student model, we sample an optimal subset of CoTs for training. Note that our CoT-based system is not static; as more powerful LLMs become publicly available, we can continually collect higher-quality CoTs to train stronger small models. This is further demonstrated by the significant improvement of \textbf{DistilQwen-ThoughtY} over \textbf{DistilQwen-ThoughtX} shown in our experiments.

\begin{table*}
\centering
\begin{small}
\begin{tabular}{lccccc}
\hline
\textbf{Model} & \textbf{Training Set Size} & \textbf{AIME2024} & \textbf{MATH-500} & \textbf{GPQA Diamond}  & \textbf{LiveCodeBench V2}\\
\hline
Qwen2.5-3B-Instruct & - & 6.7 & 62.6 & 32.8 & 11.3\\
\bf DistilQwen2.5-3B-R1 & 105K & \bf 16.7 & \bf 70.0 & \bf 34.3 & \bf 18.0\\
\hline
Qwen2.5-7B-Instruct & - & 10.0 & 73.6 & 33.3 & 30.7\\
OpenThinker-7B & 114K & 31.3 & 83.0 & 42.4 & 39.9\\
\bf DistilQwen2.5-7B-R1 & 105K & \bf 43.3 & \bf 88.4 & \bf 42.9 & \bf 46.4\\
\hline
Qwen2.5-14B-Instruct & - & 16.7 & 78.2 & 43.4 & 37.4\\
\bf DistilQwen2.5-14B-R1 & 105K & \bf 46.7 & \bf 90.8 & \bf 51.5 & \bf 54.4\\
\hline
Qwen2.5-32B-Instruct & - & 16.7 & 81.4 & 45.5 & 47.3\\
OpenThinker-32B & 114K & 66.0 & 90.6 & 61.6 & \bf 68.9\\
\bf DistilQwen2.5-32B-R1 & 105K & \bf 70.0 & \bf 93.8 & \bf 62.1 & 66.0\\
\hline
\end{tabular}
\end{small}
\caption{Performance comparison among \textbf{slow-thinking models} in terms of deep reasoning abilities. Note that \textbf{DistilQwen2.5-R1} and OpenThinker models leverage the same set of source reasoning problems (with a few filtered out by our verifier) and the same teacher model for training.}
\label{tab:r1}
\end{table*}

\subsection{Distilled Model Trainer}

\noindent\textbf{SFT Trainer with Curriculum Learning.}  
To effectively train our \textbf{slow-thinking} and \textbf{adaptive-thinking} models, we adopt supervised fine-tuning (SFT) enhanced with curriculum learning principles~\cite{DBLP:journals/ijcv/SovianyIRS22}. The training pipeline begins with medium-level CoT examples to ensure stable convergence and prevent overfitting. As training progresses, the curriculum gradually incorporates more challenging samples, promoting generalization and robustness across diverse scenarios. Please refer to the experimental results for additional details.

\noindent\textbf{RM Trainer with RV and CD Score Estimation.}  
Beyond evaluating CoT quality, RV and CD scores are also leveraged to train lightweight models as reward predictors, which can subsequently enhance the model's reasoning ability via reinforcement learning (RL). To this end, our reward model (RM) trainer produces two dedicated reward models, each initialized from Qwen2.5-7B-Instruct: one for RV and one for CD. This constitutes a special case of knowledge distillation (KD), as the prediction outcomes are derived from very large teacher models. This approach circumvents the need to directly employ the original teacher models as RMs during the more resource-demanding RL training process.

Although RL training is not the primary focus of this work, we briefly describe how \textbf{DistilQwen-Reward} models are integrated into RL training. We use Group Relative Policy Optimization (GRPO)~\cite{DBLP:journals/corr/abs-2501-12948} as a representative RL algorithm. Let \(f_{\text{RV}}(x)\) and \(f_{\text{CD}}(x)\) denote the predicted RV and CD scores by our models, given input \(x\). Note that the predicted scores are normalized to $[0,1]$. The corresponding RV and CD rewards are defined as follows:
\begin{equation}
R_{\text{RV}}\bigl(x\bigr) =
-\bigl|f_{\text{RV}}(x) - \operatorname{Clip}(f_{\text{RV}}(x), L_{\text{RV}}, H_{\text{RV}})\bigr|
\end{equation}
\begin{equation}
R_{\text{CD}}\bigl(x\bigr) =
-\bigl|f_{\text{CD}}(x) - \operatorname{Clip}(f_{\text{CD}}(x), L_{\text{CD}}, H_{\text{CD}})\bigr|
\end{equation}
where \((L_{\text{RV}}, H_{\text{RV}})\) and \((L_{\text{CD}}, H_{\text{CD}})\) are the respective score intervals for the output CoTs. If a score lies within its designated interval, the penalty term is zero; outside the interval, the penalty increases linearly with the distance to the nearest boundary. Combined with the conventional \emph{accuracy} and \emph{format} rewards (denoted as $R_{\text{fmt}}$ and $R_{\text{acc}}$, respectively) used in standard GRPO, the revised overall reward function is:
\begin{equation}
R=R_{\text{fmt}}+R_{\text{acc}}+\lambda_{RV}R_{\text{RV}}\bigl(x\bigr)+\lambda_{CD}R_{\text{CD}}\bigl(x\bigr)
\end{equation}
where $\lambda_{RV}$ and $\lambda_{CD}$ are tunable hyperparameters that determine the weighting. In our experiments, we further validate the effectiveness of the \textbf{DistilQwen-Reward} models relative to vanilla GRPO. We plan to continue developing RL-enhanced lightweight reasoning models in future work, which may extend beyond the scope of the KD techniques presented in this paper.

\section{Evaluation}

In this section, we present the evaluation results for all DistilQwen reasoning and reward models.

\subsection{Evaluation Benchmarks}

To comprehensively evaluate the reasoning capabilities of our models, we conduct experiments on challenging benchmarks spanning mathematics, programming, and question answering. Among these, AIME2024\footnote{\url{https://artofproblemsolving.com/wiki/index.php/2024_AIME_I}} features problems that require multi-step reasoning and intricate mathematical understanding. MATH500~\cite{DBLP:conf/nips/HendrycksBKABTS21} comprises 500 difficult mathematical problems across various domains, including algebra, geometry, number theory, and calculus. GPQA Diamond~\cite{DBLP:journals/corr/abs-2311-12022} is a complex question-answering benchmark designed to assess general-purpose reasoning and comprehension abilities. Finally, LiveCodeBench V2~\cite{DBLP:conf/iclr/JainHGLYZWSSS25} evaluates models' programming and code generation skills, comprising coding tasks that range from algorithmic challenges to practical problems.

\begin{figure}[t]
\centering
\subfigure[Task: MATH500.]{
\includegraphics[width=0.85\linewidth]{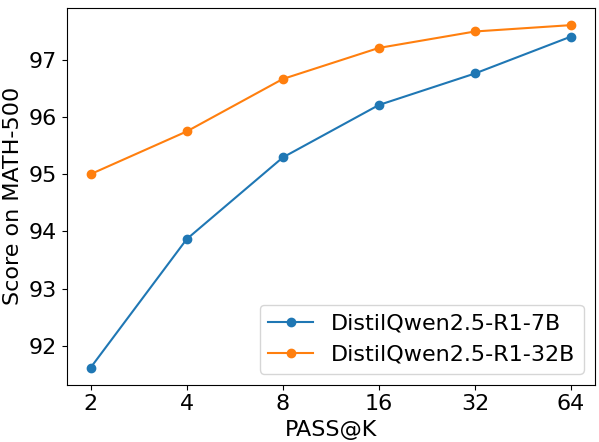}
}
\subfigure[Task: GPQA Diamond.]{
\includegraphics[width=0.85\linewidth]{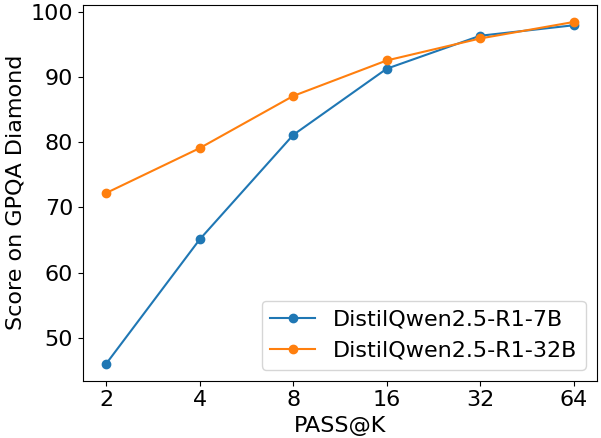}
}
\caption{Performance of \textbf{DistilQwen2.5-R1} models in terms of $Pass@K$ under multiple inference attempts.}
\label{fig:scale}
\end{figure}

\begin{table*}
\centering
\begin{small}
\begin{tabular}{lcccc}
\hline
\textbf{Model} & \textbf{AIME2024} & \textbf{MATH500} & \textbf{GPQA Diamond}  & \textbf{LiveCodeBench V2}\\
\hline
\emph{Experiments on DistilQwen-ThoughtX}\\
\hline
Qwen2.5-7B-Instruct & 10.0 & 73.6 & 33.3 & 30.7\\
\bf DistilQwen-ThoughtX-7B & \bf 56.7 & \bf 90.2 & \bf 50.0 & \bf 56.8\\ 
\hline
Qwen2.5-32B-Instruct & 16.67 & 81.4 & 45.5 & 47.3\\
\bf DistilQwen-ThoughtX-32B & \bf 80.0 & \bf 92.6 & \bf 64.0 & \bf 73.4\\
\hline
\emph{Experiments on DistilQwen-ThoughtY}\\
\hline
Qwen3-4B (thinking mode) & 73.3 & 93.2 & 54.0 & 75.7\\
\bf DistilQwen-ThoughtY-4B & \bf 76.7 & \bf 95.2 & \bf 56.1 & \bf 75.8\\ 
\hline
Qwen3-8B (thinking mode) & \bf 76.7 & 94.0 & 62.0 & 62.8\\
\bf DistilQwen-ThoughtY-8B & \bf 76.7 & \bf 94.6 & \bf 62.1 & \bf 78.1\\
\hline
Qwen3-32B (thinking mode) & 76.7 & 94.8 & \bf 65.7 & 72.2\\
\bf DistilQwen-ThoughtY-32B & \bf 90.0 & \bf 95.2 & 63.6 & \bf 76.3\\
\hline
\end{tabular}
\end{small}
\caption{Performance of \textbf{adaptive-thinking models}. Note that the Qwen2.5 models are non-reasoning models, while the Qwen3 models can act as reasoning models with their thinking modes enabled.}
\label{tab:adaptive}
\end{table*}

\subsection{Evaluation of Slow-Thinking Models}

\textbf{Slow-thinking models}, namely the \textbf{DistilQwen2.5-R1} series, encompass model scales of 3B, 7B, 14B, and 32B parameters. The source reasoning problems are taken from OpenThoughts\footnote{\url{https://huggingface.co/datasets/open-thoughts/OpenThoughts-114k}}, with CoTs generated, rewritten, and verified using DeepSeek-R1. We perform SFT training on medium-level CoTs for three epochs, followed by additional training on harder examples. After SFT, direct preference optimization (DPO)~\cite{DBLP:conf/nips/RafailovSMMEF23} is applied, yielding modest further improvements, though this is not our main focus here.

As presented in Table~\ref{tab:r1}, our proposed approach markedly enhances the reasoning abilities of existing LLMs, delivering consistent and substantial gains across multiple benchmarks compared to the original Qwen2.5 models and the OpenThinker models\footnote{\url{https://huggingface.co/open-thoughts/OpenThinker-7B}}, which are trained on the same set of reasoning problems with the same teacher model.

Additionally, we evaluate \textbf{DistilQwen2.5-R1} models using inference-time scaling, where the models generate \(k\) answers for the same question and are measured using the \(Pass@K\) metric. The findings indicate that increasing the number of reasoning attempts \(K\) leads to significant accuracy improvements for both models. Notably, the 7B model exhibits a steep upward trend on MATH500 and GPQA Diamond, gradually approaching the performance level of the 32B model while reducing inference computation requirements.

\subsection{Evaluation of Adaptive-Thinking Models}

In Table~\ref{tab:adaptive}, we present the evaluation results for our \textbf{adaptive-thinking models}. The \textbf{DistilQwen-ThoughtX} series, built upon Qwen2.5 models and trained with the dataset from~\cite{omnithought}, encompasses model scales of 7B and 32B. The results demonstrate substantial improvements at all scales, surpassing the \textbf{slow-thinking models}. The \textbf{DistilQwen-ThoughtY} series, initialized from Qwen3 models with reasoning modes enabled, achieves further advances. These models span three scales (4B, 8B, and 32B) and are trained on the previous series' dataset and a subset of 365K CoTs generated from DeepSeek-R1-0528 (to be released). The experiments confirm that our adaptive-thinking CoT processor and training strategies effectively enhance reasoning capabilities across model scales and tasks.

In addition, Table~\ref{tab:length} analyzes the output CoT lengths according to the difficulty of reasoning problems. The results clearly demonstrate that our \textbf{adaptive-thinking models} generate more optimal CoTs based on their understanding of the input. For example, the \textbf{adaptive-thinking models} produce shorter CoTs for simpler problems in GSM8K\footnote{\url{https://huggingface.co/datasets/openai/gsm8k}}, and generate longer CoTs for more challenging problems in MATH500 and AIME2024.

\begin{table}
\centering
\begin{small}
\begin{tabular}{lccc}
\hline
\textbf{Model} & \textbf{GSM8K} & \textbf{MATH500} & \textbf{AIME2024} \\
\hline
7B-R1 & 1223.61 & 6586.36 & 12856.23 \\
ThoughtX-7B & 834.36 & 6031.11 & 14597.96 \\
ThoughtY-8B & 844.23 & 5932.95 & 15632.85 \\
\hline
32B-R1 & 1178.92 & 6434.50 & 13583.19 \\
ThoughtX-32B & 742.04 & 5927.32 & 16387.53 \\
ThoughtY-32B & 723.18 & 5723.08 & 17231.84 \\
\hline
\end{tabular}
\end{small}
\caption{Analysis of averaged output CoT lengths of \textbf{adaptive-thinking models}. ``DistilQwen'' and ``DistilQwen2.5'' prefixes are omitted from model names.}
\label{tab:length}
\end{table}

\noindent\textbf{Further Discussion.}
The reason why the results obtained with adaptive-thinking models are significantly better than those with slow-thinking models is that adaptive-thinking models utilize substantially more training data. Adaptive-thinking models sample from over 2 million CoTs, resulting in training sets of at least 500K, while slow-thinking models train on only approximately 100K data points.

Although the slow-thinking recipe can effectively enhance a model's reasoning capabilities, it is unsuitable for customized, diverse training requirements due to its lack of quantitative CoT evaluation information. Therefore, we proposed the adaptive-thinking recipe, and we did not further increase the training data for slow-thinking models.

However, this does not imply that the slow-thinking recipe is obsolete. In scenarios where training data is inherently limited, the sampling process of the adaptive-thinking recipe would further reduce available training data. In such cases, we recommend using the slow-thinking recipe to ensure satisfactory training results.

\subsection{Evaluation of Reward Models}

Table~\ref{tab:reward} presents comparison results between the standard GRPO algorithm and GRPO augmented with our RV- and/or CD-based \textbf{distilled reward models}. We use Qwen2.5-7B-Instruct as the base model and randomly sample 10K mathematical problems from the previous training set for RL training. Specifically, we conduct RL training on the model directly, with no CoT-based SFT on mathematical problems, ensuring that models learn mathematical reasoning purely via RL. As shown, GRPO enhanced with RV/CD-based reward models consistently outperforms vanilla GRPO, corroborating our hypothesis that RV/CD scores distilled from teacher models can effectively benefit RL training, which is a promising direction for future work.\footnote{In SFT training, we utilize over 500K training samples, whereas in RL training, we use only 10K samples. Consequently, the results in Table~\ref{tab:reward} are lower than those in Tables~\ref{tab:r1} and Table~\ref{tab:adaptive}.}

\begin{table}
\centering
\begin{small}
\begin{tabular}{lcc}
\hline
\textbf{Reward Setting} & \textbf{MATH500} & \textbf{AIME2024} \\
\hline
Qwen2.5-7B-Instruct (Raw) & 73.6 & 10.0\\
Vanilla GRPO & 78.8	& 13.3\\
\hline
GRPO+RV & 79.0 & 13.3 \\
GRPO+CD & 80.8 & 16.7 \\
GRPO+RV+CD & \bf 81.4 & \bf 20.0\\
\hline
\end{tabular}
\end{small}
\caption{Performance of \textbf{distilled reward models}. RL performance is measured on a subset of mathematical problems using various reward settings. None of the models have undergone CoT-based SFT for mathematical reasoning tasks.}
\label{tab:reward}
\end{table}

\begin{figure*}[t]
\centering
\subfigure[List of model cards.]{
\includegraphics[width=0.3\linewidth]{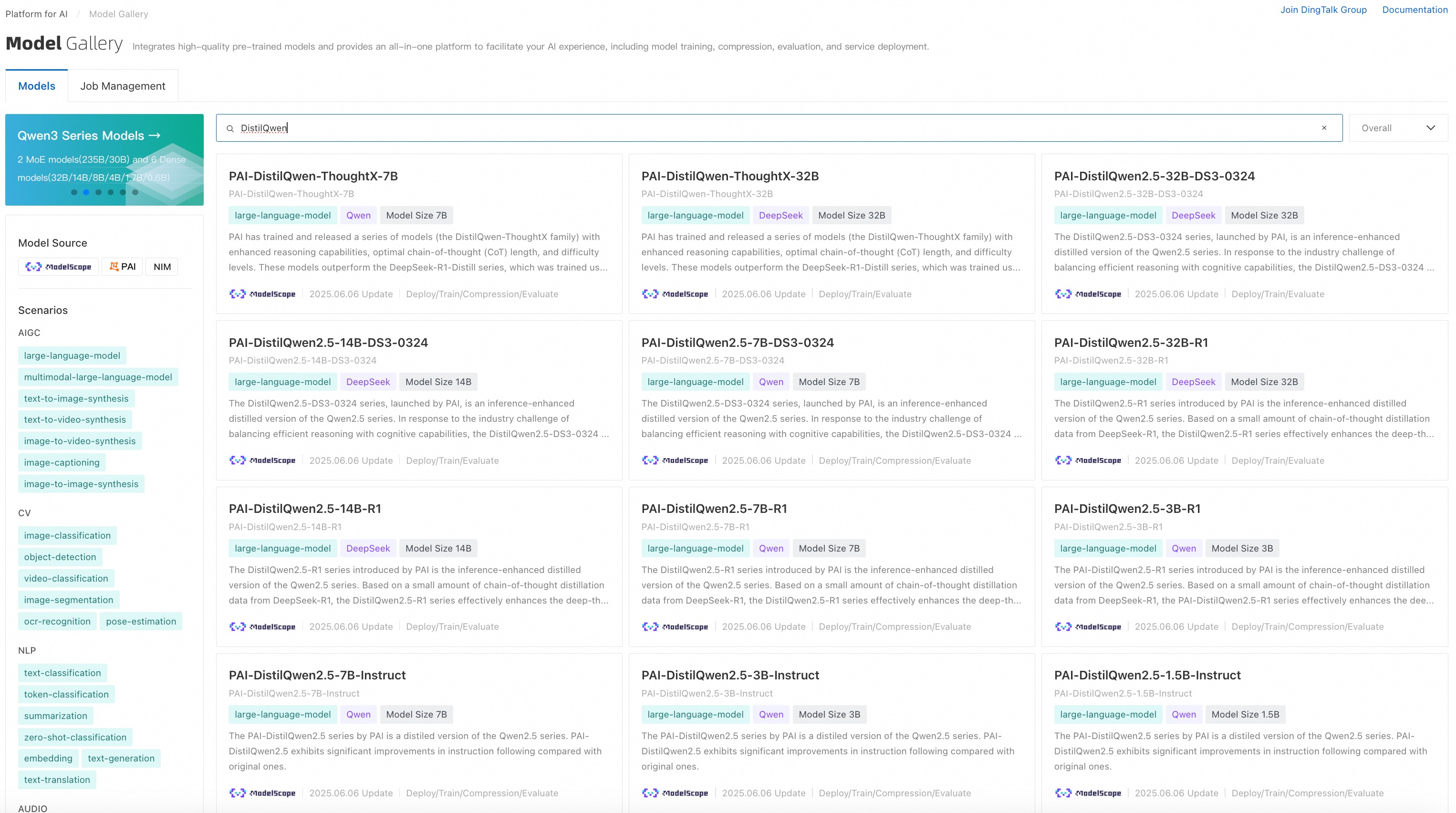}
}
\subfigure[Training panel.]{
\includegraphics[width=0.3\linewidth]{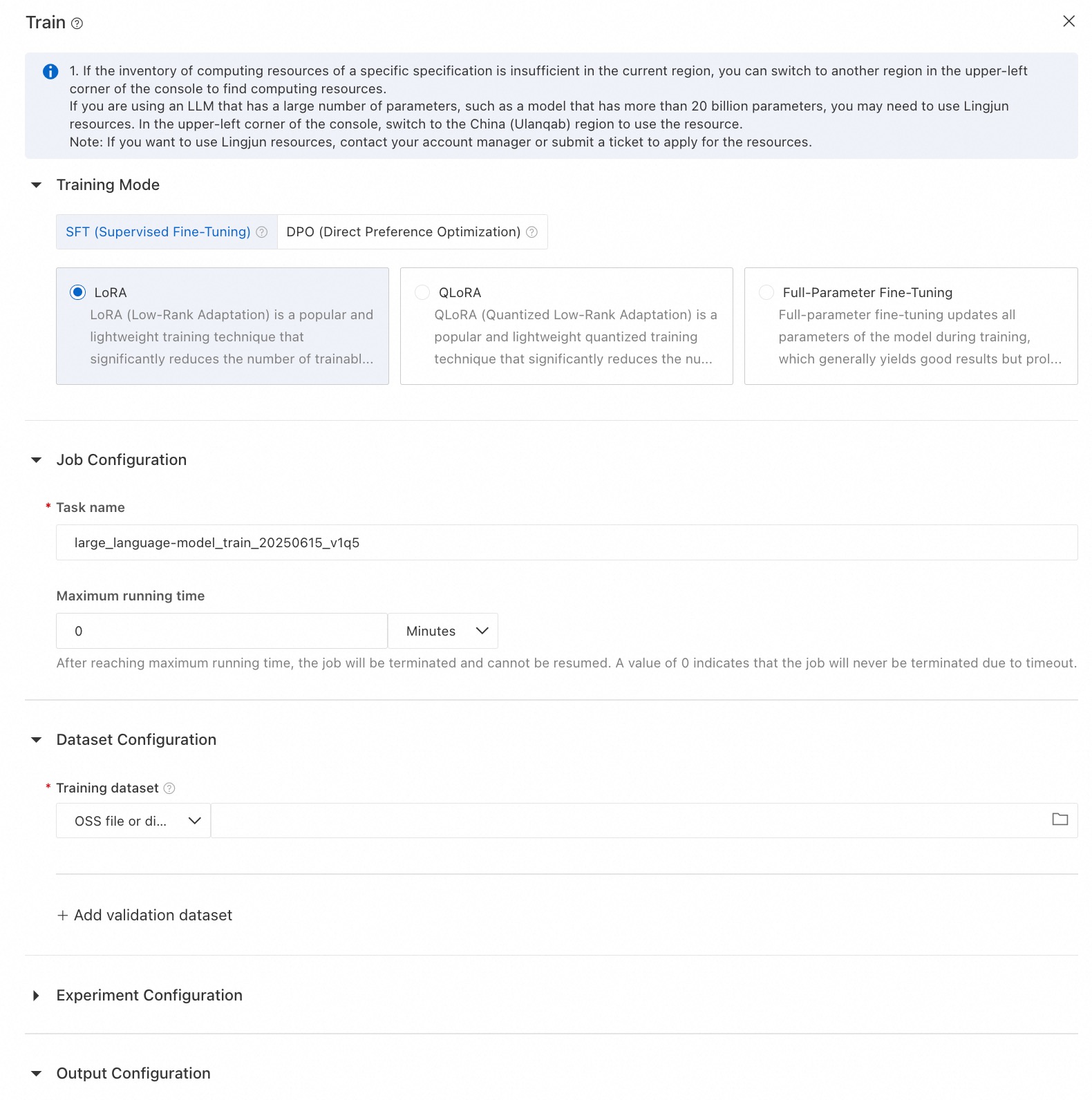}
}
\subfigure[Deployment panel.]{
\includegraphics[width=0.3\linewidth]{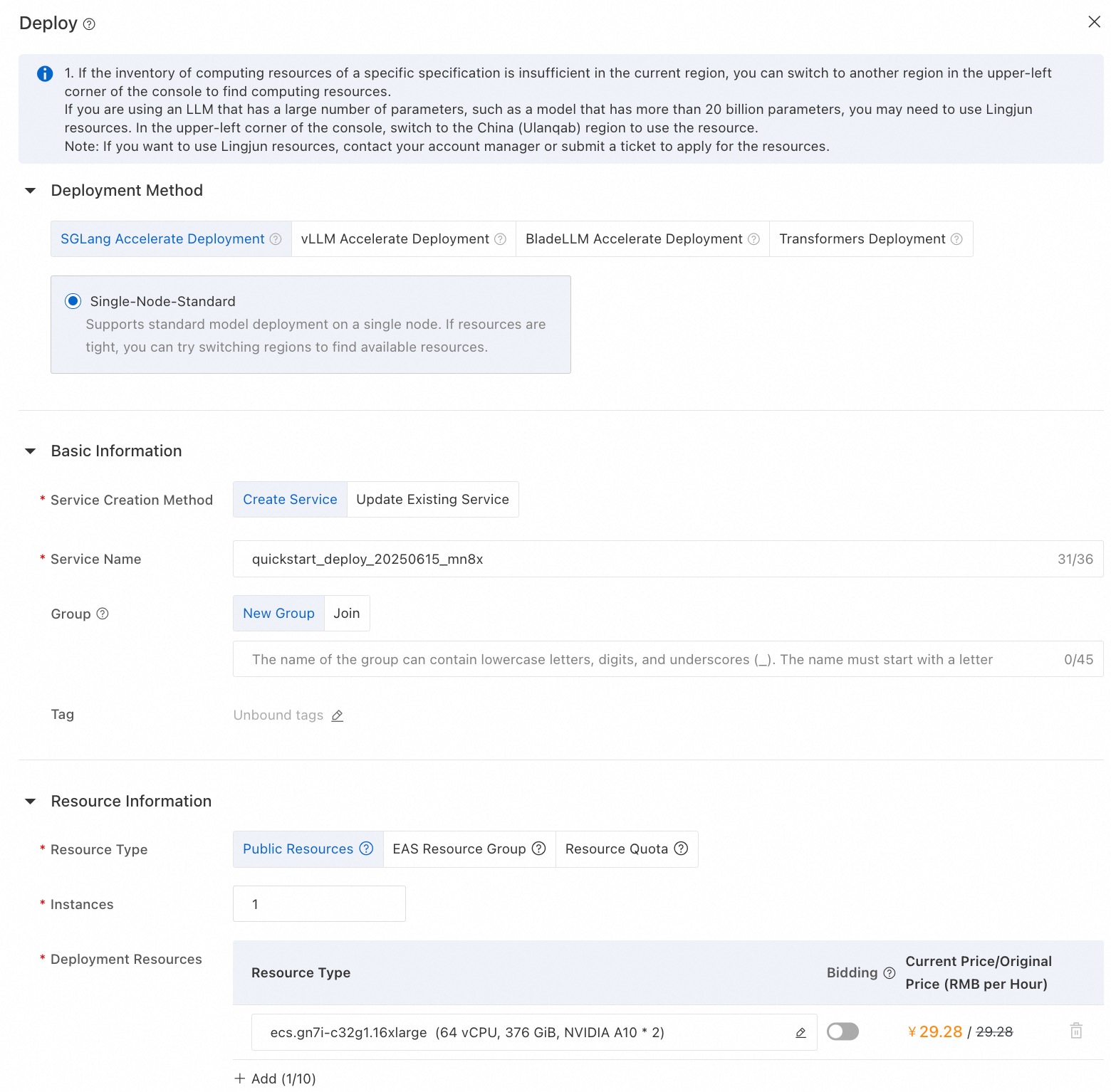}
}
\caption{Snapshots of the integration of DistilQwen reasoning models with the AI platform.}
\label{fig:mg}
\end{figure*}

\section{Industrial Solutions}

In addition to releasing all our DistilQwen reasoning models to the open-source community for use in various local environments, we have integrated these models into the Alibaba Cloud PAI (Platform for Artificial Intelligence) platform that supports the entire lifecycle of LLM usage, including training, evaluation, compression, and deployment. Snapshots of model cards as well as the training and deployment panels are shown in Figure~\ref{fig:mg}.

By embedding the DistilQwen reasoning models into AI platforms, businesses can leverage and further adapt these specialized capabilities to support real-world applications, including but not limited to decision making, code generation, problem solving, and multi-agent systems. The platform's deployment functionalities enable seamless integration of these models into existing systems via RESTful APIs compatible with the OpenAI format, thereby facilitating easier and more efficient usage.

\section{Conclusion and Future Work}

In this work, we have expanded the DistilQwen model collection by introducing four model series tailored to address diverse reasoning requirements. Our \textbf{slow-thinking models} prioritize accuracy for demanding tasks, while the \textbf{adaptive-thinking models} dynamically optimize reasoning strategies to balance efficiency and performance. In addition, the \textbf{distilled reward models} facilitate further enhancement through reinforcement learning based on distilled knowledge. Extensive evaluations demonstrate that our models achieve a favorable trade-off between inference efficiency and reasoning capability. Furthermore, by integrating these models into a scalable AI platform, we provide practical tools that effectively support industry practitioners in model training and deployment.

\section*{Limitations}

Although the proposed DistilQwen reasoning models perform well across several benchmarks, their effectiveness may vary in highly specific or dynamic real-world contexts where benchmarks do not fully capture operational complexity. Furthermore, the reinforcement learning framework supported by the distilled reward models depends on the quality of the distilled knowledge, which may propagate biases or errors inherent in the teacher models. We suggest that future research focus on addressing these limitations and further improving model adaptability and precision in real-world applications.

\section*{Ethical Considerations}

The development and deployment of the DistilQwen reasoning models requires careful consideration of ethical implications. It is essential to address potential biases introduced by the training data, as such biases may influence reasoning outcomes. Since our models are intended for research and industrial use, issues of data privacy, security, and compliance with relevant regulations should be rigorously addressed during practical implementation.

\section*{Acknowledgments}

This work was supported by Alibaba Research Intern Program.


\begin{thebibliography}{37}
\expandafter\ifx\csname natexlab\endcsname\relax\def\natexlab#1{#1}\fi

\bibitem[{Cai et~al.(2025{\natexlab{a}})Cai, Wang, Yan, Huang, and Fang}]{omnithought}
Wenrui Cai, Chengyu Wang, Junbing Yan, Jun Huang, and Xiangzhong Fang. 2025{\natexlab{a}}.
\newblock \href {https://arxiv.org/abs/2505.10937} {Reasoning with omnithought: A large cot dataset with verbosity and cognitive difficulty annotations}.
\newblock \emph{CoRR}, abs/2505.10937.

\bibitem[{Cai et~al.(2025{\natexlab{b}})Cai, Wang, Yan, Huang, and Fang}]{DBLP:journals/corr/abs-2504-09802}
Wenrui Cai, Chengyu Wang, Junbing Yan, Jun Huang, and Xiangzhong Fang. 2025{\natexlab{b}}.
\newblock \href {https://doi.org/10.48550/ARXIV.2504.09802} {Training small reasoning llms with cognitive preference alignment}.
\newblock \emph{CoRR}, abs/2504.09802.

\bibitem[{Cha et~al.(2025)Cha, Park, Choi, Ryu, and Seo}]{DBLP:conf/chi/ChaPCRS25}
Seungeon Cha, Jinseok Park, Hojin Choi, Hokyoung Ryu, and Kyoungwon Seo. 2025.
\newblock \href {https://doi.org/10.1145/3706599.3720111} {{CLONE:} synthetic guideline-based clinical reasoning with large language models for early diagnosis of mild cognitive impairment}.
\newblock In \emph{Proceedings of the Extended Abstracts of the {CHI} Conference on Human Factors in Computing Systems}, pages 122:1--122:14. {ACM}.

\bibitem[{Chen et~al.(2025)Chen, Sun, Guo, Zhang, Chen, Sun, Su, Pan, Klakow, Li, and Shen}]{DBLP:journals/corr/abs-2502-18001}
Xinghao Chen, Zhijing Sun, Wenjin Guo, Miaoran Zhang, Yanjun Chen, Yirong Sun, Hui Su, Yijie Pan, Dietrich Klakow, Wenjie Li, and Xiaoyu Shen. 2025.
\newblock \href {https://doi.org/10.48550/ARXIV.2502.18001} {Unveiling the key factors for distilling chain-of-thought reasoning}.
\newblock \emph{CoRR}, abs/2502.18001.

\bibitem[{DeepSeek{-}AI(2025)}]{DBLP:journals/corr/abs-2501-12948}
DeepSeek{-}AI. 2025.
\newblock \href {https://doi.org/10.48550/ARXIV.2501.12948} {Deepseek-r1: Incentivizing reasoning capability in llms via reinforcement learning}.
\newblock \emph{CoRR}, abs/2501.12948.

\bibitem[{Gu et~al.(2024)Gu, Dong, Wei, and Huang}]{DBLP:conf/iclr/Gu0WH24}
Yuxian Gu, Li~Dong, Furu Wei, and Minlie Huang. 2024.
\newblock \href {https://openreview.net/forum?id=5h0qf7IBZZ} {Minillm: Knowledge distillation of large language models}.
\newblock In \emph{The Twelfth International Conference on Learning Representations}. OpenReview.net.

\bibitem[{He et~al.(2025)He, Liang, Xu, Liu, Chen, Wang, Song, Yu, Liang, Wang, Zhang, Wang, Tu, Mi, and Yu}]{DBLP:journals/corr/abs-2504-11456}
Zhiwei He, Tian Liang, Jiahao Xu, Qiuzhi Liu, Xingyu Chen, Yue Wang, Linfeng Song, Dian Yu, Zhenwen Liang, Wenxuan Wang, Zhuosheng Zhang, Rui Wang, Zhaopeng Tu, Haitao Mi, and Dong Yu. 2025.
\newblock \href {https://doi.org/10.48550/ARXIV.2504.11456} {Deepmath-103k: {A} large-scale, challenging, decontaminated, and verifiable mathematical dataset for advancing reasoning}.
\newblock \emph{CoRR}, abs/2504.11456.

\bibitem[{Hendrycks et~al.(2021)Hendrycks, Burns, Kadavath, Arora, Basart, Tang, Song, and Steinhardt}]{DBLP:conf/nips/HendrycksBKABTS21}
Dan Hendrycks, Collin Burns, Saurav Kadavath, Akul Arora, Steven Basart, Eric Tang, Dawn Song, and Jacob Steinhardt. 2021.
\newblock \href {https://datasets-benchmarks-proceedings.neurips.cc/paper/2021/hash/be83ab3ecd0db773eb2dc1b0a17836a1-Abstract-round2.html} {Measuring mathematical problem solving with the {MATH} dataset}.
\newblock In \emph{Proceedings of the Neural Information Processing Systems Track on Datasets and Benchmarks}.

\bibitem[{Hinton et~al.(2015)Hinton, Vinyals, and Dean}]{DBLP:journals/corr/HintonVD15}
Geoffrey~E. Hinton, Oriol Vinyals, and Jeffrey Dean. 2015.
\newblock \href {http://arxiv.org/abs/1503.02531} {Distilling the knowledge in a neural network}.
\newblock \emph{CoRR}, abs/1503.02531.

\bibitem[{Hsieh et~al.(2023)Hsieh, Li, Yeh, Nakhost, Fujii, Ratner, Krishna, Lee, and Pfister}]{DBLP:conf/acl/HsiehLYNFRKLP23}
Cheng{-}Yu Hsieh, Chun{-}Liang Li, Chih{-}Kuan Yeh, Hootan Nakhost, Yasuhisa Fujii, Alex Ratner, Ranjay Krishna, Chen{-}Yu Lee, and Tomas Pfister. 2023.
\newblock \href {https://doi.org/10.18653/v1/2023.findings-acl.507} {Distilling step-by-step! outperforming larger language models with less training data and smaller model sizes}.
\newblock In \emph{Findings of the Association for Computational Linguistics: {ACL} 2023}, pages 8003--8017. Association for Computational Linguistics.

\bibitem[{Jain et~al.(2025)Jain, Han, Gu, Li, Yan, Zhang, Wang, Solar{-}Lezama, Sen, and Stoica}]{DBLP:conf/iclr/JainHGLYZWSSS25}
Naman Jain, King Han, Alex Gu, Wen{-}Ding Li, Fanjia Yan, Tianjun Zhang, Sida Wang, Armando Solar{-}Lezama, Koushik Sen, and Ion Stoica. 2025.
\newblock \href {https://openreview.net/forum?id=chfJJYC3iL} {Livecodebench: Holistic and contamination free evaluation of large language models for code}.
\newblock In \emph{The Thirteenth International Conference on Learning Representations}. OpenReview.net.

\bibitem[{Jiao et~al.(2020)Jiao, Yin, Shang, Jiang, Chen, Li, Wang, and Liu}]{DBLP:conf/emnlp/JiaoYSJCL0L20}
Xiaoqi Jiao, Yichun Yin, Lifeng Shang, Xin Jiang, Xiao Chen, Linlin Li, Fang Wang, and Qun Liu. 2020.
\newblock \href {https://doi.org/10.18653/v1/2020.findings-emnlp.372} {Tinybert: Distilling {BERT} for natural language understanding}.
\newblock In \emph{Findings of the Association for Computational Linguistics: {EMNLP} 2020}, volume {EMNLP} 2020 of \emph{Findings of {ACL}}, pages 4163--4174. Association for Computational Linguistics.

\bibitem[{Kim et~al.(2024)Kim, Jung, and Koo}]{DBLP:conf/naacl/KimJK24}
Minju Kim, Haein Jung, and Myoung{-}Wan Koo. 2024.
\newblock \href {https://doi.org/10.18653/v1/2024.findings-naacl.69} {{SELF-EXPERTISE:} knowledge-based instruction dataset augmentation for a legal expert language model}.
\newblock In \emph{Findings of the Association for Computational Linguistics: {NAACL} 2024}, pages 1098--1112. Association for Computational Linguistics.

\bibitem[{Li et~al.(2025)Li, Zhou, and Song}]{DBLP:conf/coling/LiZS25}
Minchong Li, Feng Zhou, and Xiaohui Song. 2025.
\newblock \href {https://aclanthology.org/2025.coling-main.78/} {Bild: Bi-directional logits difference loss for large language model distillation}.
\newblock In \emph{Proceedings of the 31st International Conference on Computational Linguistics}, pages 1168--1182. Association for Computational Linguistics.

\bibitem[{Li et~al.(2024)Li, Chen, Chen, He, Gu, and Zhou}]{DBLP:conf/acl/LiCCHGZ24}
Ming Li, Lichang Chen, Jiuhai Chen, Shwai He, Jiuxiang Gu, and Tianyi Zhou. 2024.
\newblock \href {https://doi.org/10.18653/v1/2024.findings-acl.958} {Selective reflection-tuning: Student-selected data recycling for {LLM} instruction-tuning}.
\newblock In \emph{Findings of the Association for Computational Linguistics, {ACL} 2024}, pages 16189--16211. Association for Computational Linguistics.

\bibitem[{Lou et~al.(2024)Lou, Zhang, Xie, Sun, Ahn, Xu, Su, and Yin}]{DBLP:conf/iclr/Lou0XSAX0024}
Renze Lou, Kai Zhang, Jian Xie, Yuxuan Sun, Janice Ahn, Hanzi Xu, Yu~Su, and Wenpeng Yin. 2024.
\newblock \href {https://openreview.net/forum?id=1vrS1zwekw} {{MUFFIN:} curating multi-faceted instructions for improving instruction following}.
\newblock In \emph{The Twelfth International Conference on Learning Representations}. OpenReview.net.

\bibitem[{Ma et~al.(2025)Ma, Chen, Zhang, Wu, and Ding}]{DBLP:conf/aaai/MaCZ0D25}
Weijian Ma, Ruoxin Chen, Ke{-}Yue Zhang, Shuang Wu, and Shouhong Ding. 2025.
\newblock \href {https://doi.org/10.1609/aaai.v39i6.32640} {Instruct where the model fails: Generative data augmentation via guided self-contrastive fine-tuning}.
\newblock In \emph{AAAI-25, Sponsored by the Association for the Advancement of Artificial Intelligence}, pages 5991--5999. {AAAI} Press.

\bibitem[{Pan et~al.(2021)Pan, Wang, Qiu, Zhang, Li, and Huang}]{DBLP:conf/acl/Pan0QZLH20}
Haojie Pan, Chengyu Wang, Minghui Qiu, Yichang Zhang, Yaliang Li, and Jun Huang. 2021.
\newblock \href {https://doi.org/10.18653/v1/2021.acl-long.236} {Meta-kd: {A} meta knowledge distillation framework for language model compression across domains}.
\newblock In \emph{Proceedings of the 59th Annual Meeting of the Association for Computational Linguistics and the 11th International Joint Conference on Natural Language Processing}, pages 3026--3036. Association for Computational Linguistics.

\bibitem[{Rafailov et~al.(2023)Rafailov, Sharma, Mitchell, Manning, Ermon, and Finn}]{DBLP:conf/nips/RafailovSMMEF23}
Rafael Rafailov, Archit Sharma, Eric Mitchell, Christopher~D. Manning, Stefano Ermon, and Chelsea Finn. 2023.
\newblock \href {http://papers.nips.cc/paper\_files/paper/2023/hash/a85b405ed65c6477a4fe8302b5e06ce7-Abstract-Conference.html} {Direct preference optimization: Your language model is secretly a reward model}.
\newblock In \emph{Advances in Neural Information Processing Systems 36: Annual Conference on Neural Information Processing Systems 2023}.

\bibitem[{Ranaldi and Freitas(2024)}]{DBLP:conf/emnlp/RanaldiF24}
Leonardo Ranaldi and Andr{\'{e}} Freitas. 2024.
\newblock \href {https://aclanthology.org/2024.emnlp-main.139} {Self-refine instruction-tuning for aligning reasoning in language models}.
\newblock In \emph{Proceedings of the 2024 Conference on Empirical Methods in Natural Language Processing}, pages 2325--2347. Association for Computational Linguistics.

\bibitem[{Rein et~al.(2023)Rein, Hou, Stickland, Petty, Pang, Dirani, Michael, and Bowman}]{DBLP:journals/corr/abs-2311-12022}
David Rein, Betty~Li Hou, Asa~Cooper Stickland, Jackson Petty, Richard~Yuanzhe Pang, Julien Dirani, Julian Michael, and Samuel~R. Bowman. 2023.
\newblock \href {https://doi.org/10.48550/ARXIV.2311.12022} {{GPQA:} {A} graduate-level google-proof q{\&}a benchmark}.
\newblock \emph{CoRR}, abs/2311.12022.

\bibitem[{Sanh et~al.(2019)Sanh, Debut, Chaumond, and Wolf}]{DBLP:journals/corr/abs-1910-01108}
Victor Sanh, Lysandre Debut, Julien Chaumond, and Thomas Wolf. 2019.
\newblock \href {http://arxiv.org/abs/1910.01108} {Distilbert, a distilled version of {BERT:} smaller, faster, cheaper and lighter}.
\newblock \emph{CoRR}, abs/1910.01108.

\bibitem[{Soviany et~al.(2022)Soviany, Ionescu, Rota, and Sebe}]{DBLP:journals/ijcv/SovianyIRS22}
Petru Soviany, Radu~Tudor Ionescu, Paolo Rota, and Nicu Sebe. 2022.
\newblock \href {https://doi.org/10.1007/S11263-022-01611-X} {Curriculum learning: {A} survey}.
\newblock \emph{Int. J. Comput. Vis.}, 130(6):1526--1565.

\bibitem[{Sui et~al.(2025)Sui, Chuang, Wang, Zhang, Zhang, Yuan, Liu, Wen, Zhong, Chen, and Hu}]{DBLP:journals/corr/abs-2503-16419}
Yang Sui, Yu{-}Neng Chuang, Guanchu Wang, Jiamu Zhang, Tianyi Zhang, Jiayi Yuan, Hongyi Liu, Andrew Wen, Shaochen Zhong, Hanjie Chen, and Xia~Ben Hu. 2025.
\newblock \href {https://doi.org/10.48550/ARXIV.2503.16419} {Stop overthinking: {A} survey on efficient reasoning for large language models}.
\newblock \emph{CoRR}, abs/2503.16419.

\bibitem[{Sun et~al.(2020)Sun, Yu, Song, Liu, Yang, and Zhou}]{DBLP:conf/acl/SunYSLYZ20}
Zhiqing Sun, Hongkun Yu, Xiaodan Song, Renjie Liu, Yiming Yang, and Denny Zhou. 2020.
\newblock \href {https://doi.org/10.18653/v1/2020.acl-main.195} {Mobilebert: a compact task-agnostic {BERT} for resource-limited devices}.
\newblock In \emph{Proceedings of the 58th Annual Meeting of the Association for Computational Linguistics}. Association for Computational Linguistics.

\bibitem[{Tian et~al.(2025)Tian, Han, Chen, Wang, and Chawla}]{DBLP:conf/wsdm/0001HC0C25}
Yijun Tian, Yikun Han, Xiusi Chen, Wei Wang, and Nitesh~V. Chawla. 2025.
\newblock \href {https://doi.org/10.1145/3701551.3703577} {Beyond answers: Transferring reasoning capabilities to smaller llms using multi-teacher knowledge distillation}.
\newblock In \emph{Proceedings of the Eighteenth {ACM} International Conference on Web Search and Data Mining}, pages 251--260. {ACM}.

\bibitem[{Vrdoljak et~al.(2025)Vrdoljak, Boban, Males, Skrabic, Kumric, Ottosen, Clemencau, Bozic, and V{\"{o}}lker}]{DBLP:journals/cbm/VrdoljakBMSKOCBV25}
Josip Vrdoljak, Zvonimir Boban, Ivan Males, Roko Skrabic, Marko Kumric, Anna Ottosen, Alexander Clemencau, Josko Bozic, and Sebastian V{\"{o}}lker. 2025.
\newblock \href {https://doi.org/10.1016/j.compbiomed.2025.110351} {Evaluating large language and large reasoning models as decision support tools in emergency internal medicine}.
\newblock \emph{Comput. Biol. Medicine}, 192:110351.

\bibitem[{Wang et~al.(2025{\natexlab{a}})Wang, Yan, Cai, Yue, and Huang}]{DBLP:journals/corr/abs-2505-20888}
Chengyu Wang, Junbing Yan, Wenrui Cai, Yuanhao Yue, and Jun Huang. 2025{\natexlab{a}}.
\newblock \href {https://doi.org/10.48550/ARXIV.2505.20888} {Easydistill: {A} comprehensive toolkit for effective knowledge distillation of large language models}.
\newblock \emph{CoRR}, abs/2505.20888.

\bibitem[{Wang et~al.(2025{\natexlab{b}})Wang, Yan, Yue, and Huang}]{DBLP:journals/corr/abs-2504-15027}
Chengyu Wang, Junbing Yan, Yuanhao Yue, and Jun Huang. 2025{\natexlab{b}}.
\newblock \href {https://doi.org/10.48550/ARXIV.2504.15027} {Distilqwen2.5: Industrial practices of training distilled open lightweight language models}.
\newblock \emph{CoRR}, abs/2504.15027.

\bibitem[{Wang et~al.(2025{\natexlab{c}})Wang, Zhang, Hong, and Huang}]{DBLP:journals/corr/abs-2504-09100}
Chengyu Wang, Taolin Zhang, Richang Hong, and Jun Huang. 2025{\natexlab{c}}.
\newblock \href {https://doi.org/10.48550/ARXIV.2504.09100} {A short survey on small reasoning models: Training, inference, applications and research directions}.
\newblock \emph{CoRR}, abs/2504.09100.

\bibitem[{Wen et~al.(2023)Wen, Li, Du, and Mou}]{DBLP:conf/acl/Wen0DM23}
Yuqiao Wen, Zichao Li, Wenyu Du, and Lili Mou. 2023.
\newblock \href {https://doi.org/10.18653/v1/2023.acl-long.605} {f-divergence minimization for sequence-level knowledge distillation}.
\newblock In \emph{Proceedings of the 61st Annual Meeting of the Association for Computational Linguistics}, pages 10817--10834. Association for Computational Linguistics.

\bibitem[{Wu et~al.(2025)Wu, Tao, Wang, Yang, Zhao, and Wong}]{DBLP:conf/coling/WuTWY0W25}
Taiqiang Wu, Chaofan Tao, Jiahao Wang, Runming Yang, Zhe Zhao, and Ngai Wong. 2025.
\newblock \href {https://aclanthology.org/2025.coling-main.383/} {Rethinking kullback-leibler divergence in knowledge distillation for large language models}.
\newblock In \emph{Proceedings of the 31st International Conference on Computational Linguistics}, pages 5737--5755. Association for Computational Linguistics.

\bibitem[{Xu et~al.(2024)Xu, Li, Tao, Shen, Cheng, Li, Xu, Tao, and Zhou}]{DBLP:journals/corr/abs-2402-13116}
Xiaohan Xu, Ming Li, Chongyang Tao, Tao Shen, Reynold Cheng, Jinyang Li, Can Xu, Dacheng Tao, and Tianyi Zhou. 2024.
\newblock \href {https://doi.org/10.48550/ARXIV.2402.13116} {A survey on knowledge distillation of large language models}.
\newblock \emph{CoRR}, abs/2402.13116.

\bibitem[{Yu et~al.(2025)Yu, Wu, Chen, Zhang, Mei, Huang, Tan, Du, Liu, and Zhu}]{DBLP:journals/corr/abs-2504-11919}
Qianjin Yu, Keyu Wu, Zihan Chen, Chushu Zhang, Manlin Mei, Lingjun Huang, Fang Tan, Yongsheng Du, Kunlin Liu, and Yurui Zhu. 2025.
\newblock \href {https://doi.org/10.48550/ARXIV.2504.11919} {Rethinking the generation of high-quality cot data from the perspective of llm-adaptive question difficulty grading}.
\newblock \emph{CoRR}, abs/2504.11919.

\bibitem[{Yue et~al.(2024)Yue, Wang, Huang, and Wang}]{DBLP:conf/emnlp/YueWHW24}
Yuanhao Yue, Chengyu Wang, Jun Huang, and Peng Wang. 2024.
\newblock \href {https://aclanthology.org/2024.findings-emnlp.350} {Distilling instruction-following abilities of large language models with task-aware curriculum planning}.
\newblock In \emph{Findings of the Association for Computational Linguistics: {EMNLP} 2024}, pages 6030--6054. Association for Computational Linguistics.

\bibitem[{Zhang et~al.(2025)Zhang, Zhang, Li, Li, Cheng, Chen, Wei, Ma, Wang, and Xiao}]{DBLP:journals/corr/abs-2502-12744}
Yong Zhang, Bingyuan Zhang, Zhitao Li, Ming Li, Ning Cheng, Minchuan Chen, Tao Wei, Jun Ma, Shaojun Wang, and Jing Xiao. 2025.
\newblock \href {https://doi.org/10.48550/ARXIV.2502.12744} {Self-enhanced reasoning training: Activating latent reasoning in small models for enhanced reasoning distillation}.
\newblock \emph{CoRR}, abs/2502.12744.

\bibitem[{Zhong et~al.(2025)Zhong, Huang, Wu, Luo, and Yu}]{DBLP:journals/kbs/ZhongHWLY25}
Wuchang Zhong, Jinglin Huang, Maoqiang Wu, Weinan Luo, and Rong Yu. 2025.
\newblock \href {https://doi.org/10.1016/j.knosys.2025.113630} {Large language model based system with causal inference and chain-of-thoughts reasoning for traffic scene risk assessment}.
\newblock \emph{Knowl. Based Syst.}, 319:113630.

\end{thebibliography}
\end{document}